\documentclass[conference,preprint]{IEEEtran}
\IEEEoverridecommandlockouts
\usepackage{cite}
\usepackage{amsmath,amssymb,amsfonts}
\usepackage{algorithmic}
\usepackage{graphicx}
\usepackage{textcomp}
\usepackage{xcolor}
\usepackage{booktabs} 
\usepackage[autostyle]{csquotes}
\def\BibTeX{{\rm B\kern-.05em{\sc i\kern-.025em b}\kern-.08em
    T\kern-.1667em\lower.7ex\hbox{E}\kern-.125emX}}

\usepackage{tikz}
\usepackage{lipsum}

\makeatletter
\newcommand{\linebreakand}{%
  \end{@IEEEauthorhalign}
  \hfill\mbox{}\par
  \mbox{}\hfill\begin{@IEEEauthorhalign}
}
\makeatother

\newcommand\copyrighttext{%
  \footnotesize
  © 2026 IEEE. Personal use of this material is permitted.
  Permission from IEEE must be obtained for all other uses, in any
  current or future media, including reprinting/republishing this
  material for advertising or promotional purposes, creating new
  collective works, for resale or redistribution to servers or lists,
  or reuse of any copyrighted component of this work in other works.%
}

\newcommand\publicationnotice{%
  This work has been accepted for publication in 2026 36th Irish Signals
  and Systems Conference (ISSC). The final published
  version will be available via IEEE Xplore.%
}

\newcommand\copyrightnotice{%
  \begin{tikzpicture}[remember picture,overlay]
    \node[
      name=copyrightbox,
      anchor=south,
      yshift=30pt,
      draw,
      line width=0.2pt,
      inner xsep=4pt,
      inner ysep=3pt,
      outer sep=0pt,
      text width=\dimexpr\textwidth-8.4pt\relax,
      align=justify
    ] at (current page.south) {%
      \copyrighttext
    };

    \node[
      anchor=north,
      yshift=-3pt,
      text width=\textwidth,
      align=center,
      inner sep=0pt,
      font=\scriptsize
    ] at (copyrightbox.south) {%
      \publicationnotice
    };
  \end{tikzpicture}%
}

\begin{document}

\title{Beyond Feature Importance: A Comparative Analysis of Pattern Detection Methods in Cluster Interpretation}

\author{\IEEEauthorblockN{Benjamin Connor} 
\IEEEauthorblockA{\textit{School of EEECS} \\
\textit{Queen's University Belfast}\\
18 Malone Road, Belfast, BT9 5AF \\
bconnor02@qub.ac.uk}
\and
\IEEEauthorblockN{Anna Jurek-Loughrey }
\IEEEauthorblockA{\textit{School of EEECS} \\
\textit{Queen's University Belfast}\\
18 Malone Road, Belfast, BT9 5AF \\
a.jurek@qub.ac.uk}
\and
\IEEEauthorblockN{Lu Bai}
\IEEEauthorblockA{\textit{School of EEECS} \\
\textit{Queen's University Belfast}\\
18 Malone Road, Belfast, BT9 5AF \\
l.bai@qub.ac.uk}
\linebreakand
\IEEEauthorblockN{Muhammad Fahim}
\IEEEauthorblockA{\textit{School of EEECS} \\
\textit{Queen's University Belfast}\\
18 Malone Road, Belfast, BT9 5AF \\
m.fahim@qub.ac.uk}
}

\maketitle
\copyrightnotice
\begin{abstract}
Interpreting clustering outcomes remains a fundamental challenge in data analysis, particularly in domains such as healthcare where meaningful patterns must be extracted from high-dimensional data. While numerous explainability techniques exist, they are primarily designed to assess feature importance or provide local instance-level explanations rather than to identify structured patterns present within clusters. This work presents a comparative evaluation of commonly used post-hoc analysis methods for pattern detection in clustering results.

To enable controlled evaluation, we introduce a suite of synthetic datasets in which predefined patterns are systematically injected. Three widely used techniques are evaluated: a Random Forest surrogate model with permutation feature importance, LIME (Local Interpretable Model-agnostic Explanations), and principal component analysis. Results demonstrate that although each method can successfully recover relevant features, none consistently detects all injected pattern types. These findings highlight a critical gap between existing explainability tools and the requirements of pattern-level cluster interpretation, motivating the development of dedicated pattern detection methodologies.
\end{abstract}


\section{Introduction}
Clustering algorithms are widely used to uncover latent structure in complex datasets, particularly in high-dimensional domains such as biomedical research, population health analysis, and exploratory data mining. While clustering methods can successfully partition data into groups with similar characteristics, interpreting why a cluster exists and what patterns define it remains a major unresolved challenge~\cite{moshkovitz2020explainable,bandyapadhyay2023find}. In practice, the analytical value of clustering lies not only in assigning data points to groups but in identifying meaningful patterns that characterise those groups.

Existing approaches to cluster interpretation typically rely on post-hoc explainability techniques. Feature importance measures derived from surrogate models, local explanation frameworks such as LIME, and dimensionality reduction techniques such as principal component analysis (PCA) are frequently applied to understand cluster structure~\cite{loftus2022phenotype,marin2024integrating,luss2010clustering,zafar2019dlime}. However, these methods were not originally designed for pattern detection within clustering outcomes. Instead, they focus on identifying influential variables, explaining individual predictions, or capturing variance in reduced representations. As a result, their ability to recover structured multi-feature patterns remains unclear.

In many real-world applications, particularly in medical research, clusters are expected to represent interpretable behavioural or physiological patterns rather than isolated feature effects~\cite{yamga2023interpretable,loftus2022phenotype}. Such patterns may involve coordinated feature behaviour, including correlated variables, jointly elevated or suppressed values, or heterogeneous distributions within a cluster. Detecting these structures requires methods capable of identifying relationships among features rather than evaluating them independently.

A key obstacle in evaluating pattern detection methods is the absence of ground truth in real datasets. Without known patterns, it is difficult to assess whether an explanation method successfully captures meaningful structure or merely produces plausible interpretations. To address this limitation, we construct a controlled experimental framework based on synthetic datasets with explicitly injected patterns. Each dataset represents a single cluster generated from normally distributed features, into which predefined patterns are introduced while preserving realistic noise levels. This design enables systematic comparison between detected explanations and known underlying structure.

In this study, we evaluate three commonly used approaches for analysing clustering outcomes: Random Forest (RF) surrogate model using permutation feature importance, LIME and PCA. Each method is applied independently to synthetic clusters containing known pattern types, including high-value, low-value, correlation-based, and distribution-switching patterns. Through this evaluation, we examine not only whether relevant features are identified but whether relationships between features-and therefore patterns-are successfully recovered.

Our findings reveal that while these approaches highlight important variables, they fall short of reliably detecting multi-feature patterns, underscoring the need for dedicated pattern detection methods.

The contributions of this work are threefold:
\begin{enumerate}
    \item A formal framing of pattern detection as a distinct task from clustering and feature attribution.
    \item A controlled synthetic benchmark for evaluating pattern detection methods with known ground truth.
    \item A comparative analysis demonstrating systematic limitations of widely used interpretability techniques, motivating the need for dedicated pattern detection methodologies.
\end{enumerate}

By establishing these limitations empirically, this work lays the foundation for future research aimed at developing methods specifically designed to detect and quantify patterns within clustering results.

\section{Background and Related Work}
Clustering algorithms aim to discover latent structure in data by grouping observations according to similarity. While substantial research has focused on improving clustering algorithms themselves~\cite{jain2010data}, comparatively less attention has been devoted to interpreting the resulting clusters. In many applied domains, particularly healthcare and biomedical analytics, the primary objective is not the cluster assignment itself but the identification of patterns that characterise each cluster.

Cluster interpretation is commonly performed using summary statistics, centroid analysis, or feature ranking approaches~\cite{werner2023explainable,dunn2018cluster,loftus2022phenotype}. These techniques describe differences between clusters but often treat features independently, limiting their ability to reveal structured relationships such as correlated feature groups or heterogeneous distributions within a cluster. Consequently, researchers increasingly employ model explainability methods originally developed for supervised learning to analyse clustering outcomes.

A common strategy is to transform clustering into a supervised problem by training a classifier to predict cluster membership and then applying explainability tools to the classifier~\cite{alvarez2024comprehensive}. While practical, this approach implicitly assumes that methods designed for prediction explanation are suitable for pattern discovery - an assumption that has not been systematically validated.

\subsection{Surrogate Models and Feature Importance}

Surrogate models approximate the behaviour of a complex or non-interpretable system using a more interpretable predictive model~\cite{belle2021principles}. In the context of clustering interpretation, cluster labels are treated as targets, and a supervised model is trained to predict (one-versus-all) cluster membership. Explanations are then derived from the surrogate model.
Let the dataset be defined as $X \in \mathbb{R}^{n \times d}$, where $n$ denotes the number of samples and $d$ the number of features, with cluster labels $y \in \{0,1\}$. A surrogate model learns a mapping $f : X \rightarrow y$ that approximates the cluster decision boundary. In this work, a RF classifier is employed in a one-versus-rest formulation. Feature importance is computed using permutation importance, which measures the decrease in predictive performance when feature values are randomly permuted.

Permutation importance captures global predictive relevance and supports nonlinear relationships learned by the model. However, feature importance identifies influential variables but does not directly reveal structured multi-feature patterns.

\subsection{Local Interpretable Model-Agnostic Explanations (LIME)}

LIME is a local explanation framework designed to explain individual predictions of black-box models using locally faithful interpretable approximations~\cite{ribeiro2016should}.

Given an instance $x$, LIME approximates a complex model $f$ with a simpler interpretable model $g$ within the neighbourhood of $x$. The procedure consists of:
\begin{enumerate}
\item Choosing a specific data point $x$ we want to explain, 
\item Generating synthetic samples $S$ around instance $x$,
\item Querying the black-box model for predictions on $x$ and the samples from $S$,
\item Weighting samples in $S$ according to proximity to $x$,
\item Fitting an interpretable model on $S$,
\item Extracting feature contributions as explanation rules.
\end{enumerate}

Continuous variables are typically discretized into bins to produce human-readable rules describing feature ranges. In our evaluation, discretization is applied to all continuous features, with explanations restricted to a fixed number of features and generated using a predefined number of synthetic samples.

LIME provides interpretable local explanations and model-agnostic applicability. However, explanations are instance-specific and sensitive to sampling variability, making aggregation into cluster-level pattern descriptions challenging.

\subsection{Principal Component Analysis (PCA)}

PCA is a linear dimensionality reduction technique widely used for exploratory data analysis and structure discovery~\cite{jolliffe2016principal}.

Given a centred data matrix $X$, PCA identifies orthogonal directions that maximise variance:

\[
\max_{\|w\|=1} \mathrm{Var}(Xw).
\]

This optimisation leads to the eigenvalue decomposition of the covariance matrix

\[
\Sigma = \frac{1}{n} X^{T} X.
\]

Eigenvectors of $\Sigma$ define principal components, while eigenvalues represent explained variance. Each principal component is computed as

\[
z_k = X w_k,
\]

where $w_k$ denotes the loading vector associated with component $k$, which indicate contribution strength. High-magnitude loadings indicate higher importance. Similar signs indicate positive correlation, while opposite signs indicate negative correlation.

PCA effectively captures linear correlation structures among features. However, it assumes linear relationships and prioritises variance maximisation rather than semantic pattern identification. It does not guarantee that the resulting components correspond to meaningful or interpretable patterns.

\section{Problem Definition}

Clustering algorithms partition a dataset $X \in \mathbb{R}^{n \times d}$ into $K$ clusters
\[
\mathcal{C} = \{C_1, C_2, \dots, C_K\}, \quad \bigcup_{k=1}^K C_k = X,
\]
such that points within a cluster are more similar to each other than to points in other clusters~\cite{jain2010data}. While clustering assigns membership, it does not explain why clusters exist or which feature combinations define them~\cite{bandyapadhyay2023find}. We define the pattern detection problem as the task of identifying structured, multi-feature relationships within a given cluster that distinguish it from other clusters or from random noise.

\subsection{Definition of a Pattern}

Let $C_k$ denote a cluster and $X_{C_k} \in \mathbb{R}^{|C_k| \times d}$ its associated data. A pattern $P$ is defined as a subset of features 
\[
F_P \subseteq \{1,2,\dots,d\}
\]
and a joint feature behaviour over $X_{C_k}$ that satisfies one or more of the following properties:

\begin{enumerate}
    \item Value-based patterns: Features in $F_P$ take consistently high or low values across a majority of samples. 
    \item Correlation-based patterns: Features in $F_P$ are statistically dependent, either positively or negatively, within the cluster:
    \[
    \mathrm{Corr}(X_i, X_j) \neq 0, \quad \forall i,j \in F_P.
    \]
    \item Distributional patterns: Features in $F_P$ exhibit heterogeneous distributions, e.g., multimodal behaviour.
\end{enumerate}

\subsection{Objectives of Pattern Detection}

Given cluster $C_k$, the goal of a pattern detection method is to output a set of patterns
\[
\mathcal{P}_k = \{P_1, P_2, \dots, P_{N_k}\},
\]
where each pattern $P_i$ identifies:

\begin{itemize}
    \item The subset of features $F_{P_i}$ involved,
    \item The type of joint behaviour (value, correlation, distributional),
    \item The strength or prominence of the pattern within the cluster.
\end{itemize}

\subsection{Evaluation Context}

The problem is evaluated under a controlled synthetic framework where clusters contain known patterns of varying types (value-based, correlation-based, switch/multimodal). This allows rigorous assessment of whether existing methods (surrogate models, LIME, PCA) can correctly identify patterns discussed above.

\section{Synthetic Data Design}

In order to verify that pattern detection methods reliably identify meaningful structures within clustering results, a suite of controlled synthetic datasets was constructed. Evaluation on real-world datasets alone is insufficient for this purpose, as the true underlying patterns are typically unknown. Synthetic data enables the introduction of predefined ground-truth patterns, allowing objective assessment of whether an analysis method successfully recovers known structure.

Patterns were defined in collaboration with medical domain experts to reflect behaviours commonly observed in clinical datasets, including coordinated feature behaviour, extreme value characteristics, and heterogeneous distributions within patient subgroups. Each synthetic dataset was designed to represent the outcome of a single cluster produced by an upstream clustering algorithm, thereby isolating the task of pattern detection from clustering itself. Five categories of patterns were considered:

\begin{itemize}
    \item \textbf{High Values:} Selected features take values near the upper end of their distribution (2-3 features).
    \item \textbf{Low Values:} Selected features take values near the lower end of their distribution (2-3 features).
    \item \textbf{Positive Correlation:} Selected features exhibit positive linear correlation (2-4 features).
    \item \textbf{Negative Correlation:} Selected features exhibit negative linear correlation (2-3 features).
    \item \textbf{High-Low} Some features take values either near the upper or lower end of their distribution (2 high, 2 low).
    \item \textbf{Switch High--Low:} Selected features display heterogeneous behaviour, where approximately half of the samples take high values and the remaining half take low values (2 or 4 features).
\end{itemize}

Each pattern involves a predefined number of randomly selected features, ranging from two to four, as shown in the brackets. Each synthetic dataset consists of $n = 500$ samples and a varying number of features $d \in \{10, 20, 30, 40, 50, 60, 70, 80, 90, 100\}$. Prior to pattern injection, all features were independently generated from a normal distribution scaled to the interval $[0,1]$. This initialisation produces datasets without intrinsic structure, ensuring that any detectable patterns arise solely from the controlled injection procedure. For each dataset, a pattern type was selected from the predefined list and injected to randomly selected features. For each selected feature, $80\%$ of data points were modified to satisfy the pattern constraints and the remaining $20\%$ of samples were left unchanged to introduce controlled noise. This design simulates realistic clustering outcomes in which patterns are prominent but not perfectly expressed across all observations. A total of twelve unique pattern combinations were constructed. Each combination was generated across all feature dimensionalities, resulting in $120$ datasets.

\section{Experimental Results}
In this work we evaluate three pattern detection methods discussed earlier in the paper. These include RF surrogate model and the PCA method, which both assign importance scores to features, and the LIME method, which outputs a collection of weighted rules. For RF surrogate model and the LIME method, the predictive task was defined as one-vs-all classification, where each of the patterns is considered as an individual cluster. The results obtained by each of the three methods on our synthetic dataset are explained in the following sections. 
\subsection{RF Surrogate Model and PCA}
Tables \ref{tab:RF} and \ref{tab:PCA} summarise the results obtained by the RF surrogate models and the PCA method across all datasets. Both approaches assign importance scores to features, which are ranked to evaluate whether features participating in the true patterns are correctly prioritised. The first column lists the type of the pattern. For each pattern, the results are averaged across the different dataset dimensions. The second column reports the evaluation of the feature ranking using $top-m$ $accuracy$ score, calculated as a ratio between the number of pattern features put at the top of the ranking and the total number of pattern features. This is followed by the standard deviation in column three. The fourth column shows the average difference between the min importance values assigned to a pattern feature and the max importance value assigned to a non-pattern feature. The final column provides the standard deviation of the values reported in column 4.

\begin{table}[h!]
\centering
\caption{Summary of the results obtained for the RF surrogate model.}
\label{tab:RF}
\begin{tabular}{|c|c|c|c|c|}
\hline
\textbf{Pattern} & \textbf{Ranking Accuracy} & \textbf{St. Dev.} & \textbf{Margin} & \textbf{St. Dev.} \\ \hline
Positive 2 & 0.95 & 0.1581 & 0.0601 & 0.0488 \\ \hline
Negative 2 & 1.0 & 0.0 & 0.1193 & 0.0465 \\ \hline
High 2 & 1.0 & 0.0 & 0.3048 & 0.035 \\ \hline
Low 2 & 0.95 & 0.1581 & 0.138 & 0.1137 \\ \hline
Switch 2 & 1.0 & 0.0 & 0.2253 & 0.087 \\ \hline
Positive 3 & 0.9667 & 0.1054 & 0.0191 & 0.0288 \\ \hline
Negative 3 & 0.9667 & 0.1054 & 0.0181 & 0.0589 \\ \hline
High 3 & 0.9667 & 0.1054 & 0.0291 & 0.0391 \\ \hline
Low 3 & 1.0 & 0.0 & 0.0131 & 0.0126 \\ \hline
Positive 4 & 0.925 & 0.1687 & 0.0054 & 0.0117 \\ \hline
High-Low 4 & 0.875 & 0.1318 & 0.0012 & 0.0061 \\ \hline
Switch 4 & 0.875 & 0.1768 & 0.0144 & 0.0381 \\ \hline
\end{tabular}
\end{table}

\begin{table}[h!]
\centering
\caption{Summary of the results obtained for the PCA method.}
\label{tab:PCA}
\begin{tabular}{|c|c|c|c|c|}
\hline
\textbf{Pattern} & \textbf{Ranking Accuracy} & \textbf{St. Dev.} & \textbf{Margin} & \textbf{St. Dev.} \\ \hline
Positive 2 & 1.0 & 0.0 & 0.4424 & 0.0815 \\ \hline
Negative 2 & 1.0 & 0.0 & 0.4477 & 0.0681 \\ \hline
High 2 & 0.8 & 0.4216 & 0.089 & 0.2279 \\ \hline
Low 2 & 0.05 & 0.1581 & -0.2742 & 0.123 \\ \hline
Switch 2 & 0.8 & 0.4216 & 0.2106 & 0.2304 \\ \hline
Positive 3 & 1.0 & 0.0 & 0.4459 & 0.0345 \\ \hline
Negative 3 & 0.7 & 0.1892 & -0.1238 & 0.0815 \\ \hline
High 3 & 1.0 & 0.0 & 0.3373 & 0.0542 \\ \hline
Low 3 & 0.2667 & 0.3443 & -0.223 & 0.07 \\ \hline
Positive 4 & 1.0 & 0.0 & 0.3971 & 0.0273 \\ \hline
High-Low 4 & 0.525 & 0.1845 & -0.1504 & 0.0742 \\ \hline
Switch 4 & 1.0 & 0.0 & 0.3277 & 0.0488 \\ \hline
\end{tabular}
\end{table}
The RF-based method achieves consistently high ranking accuracy across most pattern types, typically exceeding 0.95 and often reaching perfect identification of top pattern features. This indicates that RF reliably assigns high importance to features associated with injected signals. PCA also achieves perfect or near-perfect ranking accuracy for several patterns, particularly for positive, negative, and switching structures. However, PCA exhibits substantially more variability, with performance dropping markedly for certain configurations, most notably the Low patterns. This contrast suggests that RF provides more stable feature prioritisation across pattern types, whereas PCA is sensitive to the statistical structure of the signal and performs well primarily when patterns align with dominant variance directions.

A key difference between the two methods emerges when examining the margin metric, which measures the separation between pattern and non-pattern features. RF generally produces small but positive margins, indicating that relevant features are usually ranked above irrelevant ones, albeit with limited separation. The small margins imply considerable overlap between importance scores, meaning that there is no clear threshold distinguishing pattern members from the remaining variables. In contrast, PCA produces large positive margins for some patterns (e.g., Positive and Negative patterns), suggesting strong separation when the underlying structure aligns with principal variance components. However, PCA also frequently yields negative margins, particularly for Low and mixed High–Low patterns. Negative margins indicate that non-pattern features are occasionally ranked higher than true pattern features, demonstrating instability and reduced robustness across pattern types. Thus, RF shows consistent but weak separation, while PCA exhibits stronger separation when successful but fails more dramatically when assumptions are violated.

Despite relatively high ranking accuracy, both methods share an important limitation: neither explicitly reveals the relationships among detected features. The injected patterns encode structured behaviour - correlations, shared magnitude shifts, or coordinated switching dynamics - yet both RF and PCA reduce the output to individual feature importance scores. As a result, the methods identify which features appear important but do not clarify whether these features belong to a coherent pattern or simply exhibit independent statistical influence. This limitation is particularly evident for higher-order patterns, where multiple interacting features define the signal but no explicit grouping or relational structure is recovered. Consequently, successful ranking does not necessarily correspond to successful pattern discovery.

\begin{figure}[h!]
    	\centering
    	\includegraphics[width=0.5\textwidth]{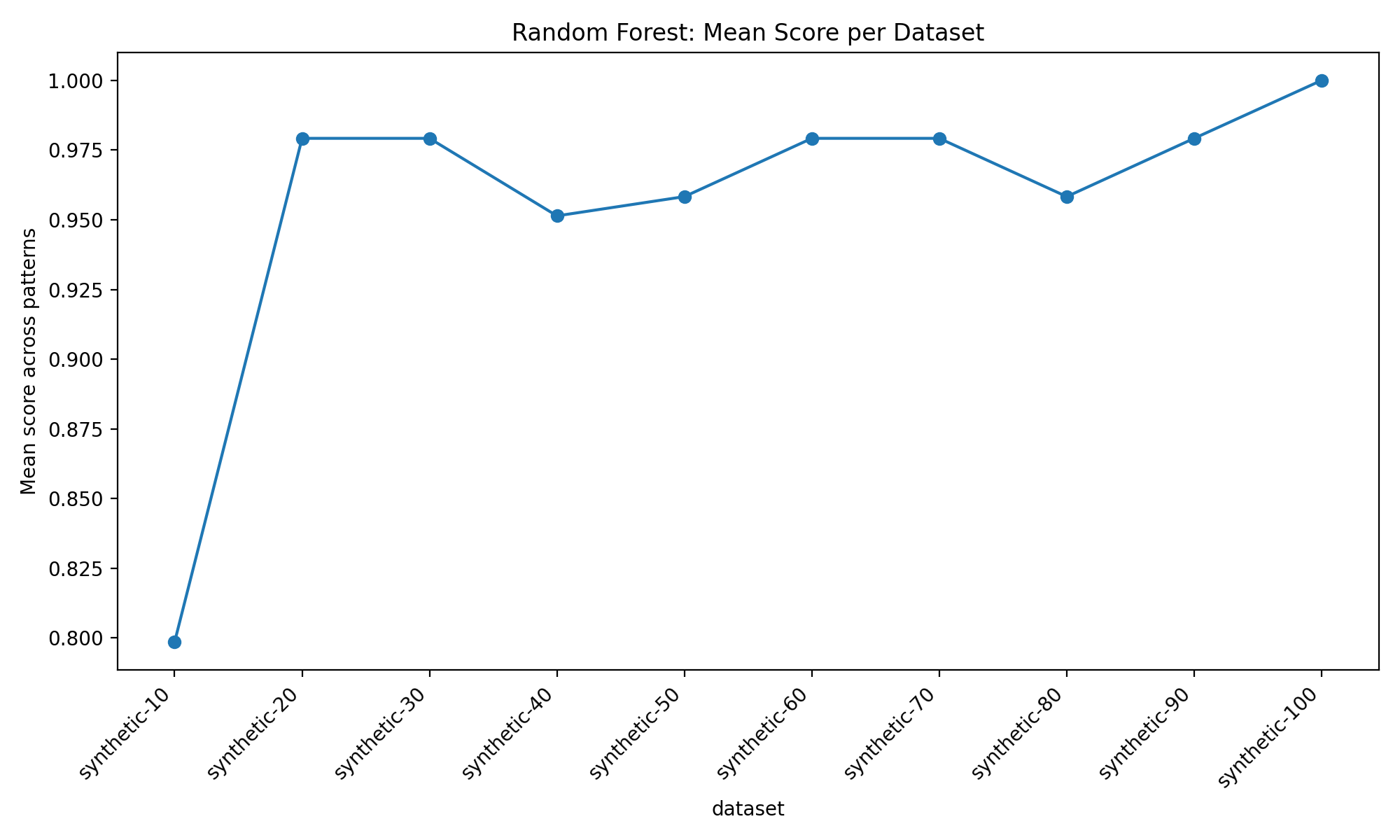}
    	\caption{Chart showing mean ranking accuracy of RF surrogate model across synthetic datasets with increasing number of dimensions.}
    \label{fig:rd_acc}
\end{figure}

\begin{figure}[h!]
    \centering
    	\includegraphics[width=0.5\textwidth]{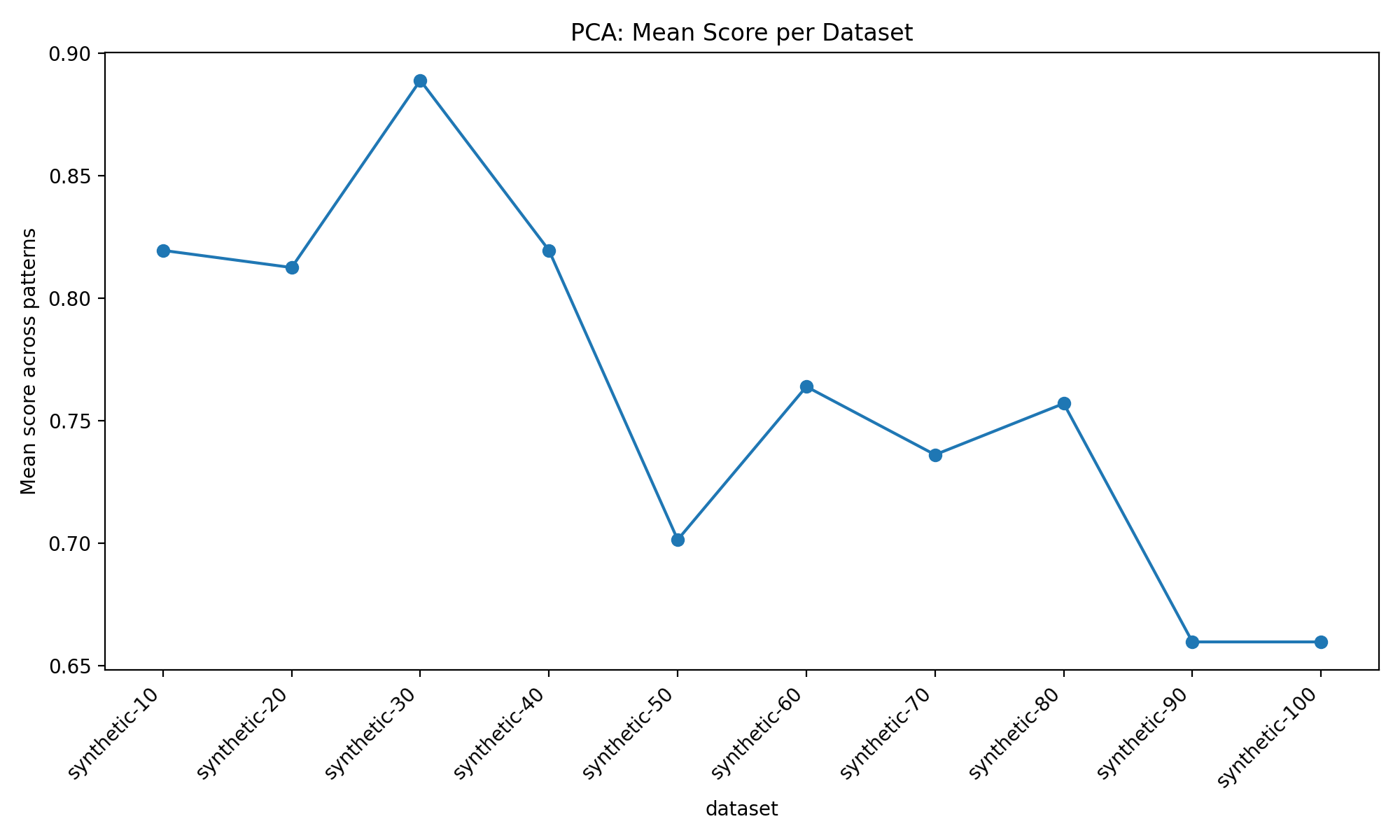}
    	\caption{Chart showing mean ranking accuracy of PCA method across synthetic datasets with increasing number of dimensions.}
\label{fig:pca_acc}
\end{figure}

We also investigate how the two methods perform across datasets of varying dimensionality. Figures \ref{fig:rd_acc} and \ref{fig:pca_acc} present the ranking accuracy as the number of features increases. The results show that the RF surrogate model maintains consistently high performance across all examined dimensionalities, indicating robustness to increasing feature-space complexity. This suggests that the RF-based importance estimation effectively isolates informative variables even when the number of irrelevant features grows. In contrast, PCA exhibits a clear decline in ranking accuracy as data dimensionality increases. This degradation can be attributed to PCA’s reliance on variance maximisation: as additional noisy or unrelated features are introduced, the dominant variance directions become increasingly influenced by non-pattern features. Consequently, the principal components capture global variance structure rather than the injected patterns, leading to reduced ability to prioritise relevant variables. These results highlight a fundamental difference between the two approaches: RF surrogate models evaluate feature importance through predictive relationships that remain relatively stable in high-dimensional settings, whereas PCA becomes progressively sensitive to variance dilution, where informative signals are overshadowed by accumulated variability from irrelevant dimensions.

%\begin{figure}[h!]
%    \centering
%    	\includegraphics[width=0.5\textwidth]{rf_mean_global_target_margin_per_dataset.png}
%    	\caption{Chart showing mean margin of RF surrogate model across synthetic datasets with increasing number of dimensions.}
%\label{fig:rd_margin}
%\end{figure}

%\begin{figure}[h!]
%    \centering
%    	\includegraphics[width=0.5\textwidth]{pca_mean_global_target_margin_per_dataset.png}
%    	\caption{Chart showing mean margin of PCA method across synthetic datasets with increasing number of dimensions.}
%\label{fig:pca_margin}
%\end{figure}

\subsection{LIME}
Table \ref{tab:LIME} summarises the explanations produced by the LIME method for datasets containing injected feature patterns, reporting the four highest-ranked rules and their associated weights for each case. Across all patterns, LIME consistently places key pattern features within the top explanatory rules, indicating that the method successfully identifies variables that influence the predictions. Importantly, no clearly irrelevant features appear among the explanations; however, individual rules typically capture only subsets of the pattern features rather than representing the full structure jointly. The interpretation of rule weights is therefore critical: positive weights indicate feature value ranges that support the prediction, whereas negative weights correspond to ranges that oppose it within the surrogate linear model fitted by LIME. Consequently, the presence of both positive and negative rules for the same feature does not imply inconsistency in the underlying model but rather reflects locally opposing contributions across neighbouring value regions. Nevertheless, this behaviour complicates interpretation, as some explanations contain adjacent intervals with opposite effects, making it difficult to derive a single coherent rule describing the pattern. Patterns defined by simple magnitude effects (e.g., consistently high or low values) can be partially inferred from the rules, since threshold conditions with strong positive weights often align with the injected feature behaviour. In contrast, correlation-based patterns cannot be directly extracted because LIME expresses explanations as independent univariate conditions and does not encode relationships between features. Similarly, switching patterns are only weakly represented: while individual high or low value conditions are detected, the coordinated opposing behaviour among feature groups is not captured as a unified rule. As a result, pattern information becomes fragmented across multiple locally valid explanations, requiring substantial manual interpretation to reconstruct the intended structure. Overall, although LIME reliably highlights influential pattern features and provides insight into value ranges supporting or opposing predictions, the combination of fragmented rules and absence of explicit feature relationships makes it difficult to recover complete and interpretable global patterns from the explanations.
\begin{table*}[!t]
\centering
\caption{Example results from LIME method for synthetic dataset with 50 features.}
\label{tab:LIME}
\begin{tabular}{|c|c|c|c|c|c|c|c|c|}
\hline
\textbf{Pattern} & \textbf{Rule 1} & \textbf{Weight} & \textbf{Rule 2} & \textbf{Weight}  & \textbf{Rule 3} & \textbf{Weight} & \textbf{Rule 4} & \textbf{Weight} \\ \hline
Positive O1 H0 & O1 \textless= -0.25 & 0.2289 & H0 \textless= -0.26 & 0.162 & -0.25 \textless O1 \textless= 0.11 & -0.1015 & 0.11 \textless O1 \textless= 0.45 & -0.0952 \\ \hline
Negative B0 V1 & V1 \textless= -0.19 & 0.1994 & B0 \textless= -0.17 & 0.1978 & -0.17 \textless B0 \textless= 0.13 & -0.0868 & 0.13 \textless B0 \textless= 0.43 & -0.0837 \\ \hline
High R0 P1 & R0 \textgreater 0.51 & 0.0959 & P0 \textgreater 0.51 & 0.0734 & -0.54 \textless R0 \textless= -0.03 & -0.055 & P0 \textless= -0.53 & 0.0339\\ \hline
Low O0 I0 & I0 \textless= -0.43 & 0.2081 & O0 \textless= -0.43 & 0.1344 & -0.43 \textless I0 \textless= 0.09 & -0.0714 & 0.09 \textless I0 \textless= 0.58 & -0.0641\\ \hline
Switch V1 G0 & G0 \textless= -0.46 & 0.1729 & -0.46 \textless G0 \textless= 0.04 & -0.0815 & V1 \textgreater 0.46 & 0.0622 & -0.19 \textless V1 \textless= 0.16 & -0.0427\\ \hline
Positive R1 V1 I1 & R1 \textless= -0.24 & 0.2151 & I1 \textless= -0.22 & 0.169 & 0.09 \textless R1 \textless= 0.43 & -0.1038 & -0.22 \textless I1 \textless= 0.14 & -0.0826\\ \hline
Negative F0 L1 B1 & F0 \textless= -0.27 & 0.2125 & B1 \textless= -0.25 & 0.212 & L1 \textless= -0.28 & 0.2035 & -0.25 \textless B1 \textless= 0.09 & -0.0892\\ \hline
High C0 B0 F0 & F0 \textgreater 0.44 & 0.0649 & C0 \textgreater 0.53 & 0.0588 & -0.27 \textless F0 \textless= 0.06 & -0.05 & -0.05 \textless C0 \textless= 0.53 & -0.0449\\ \hline
Low N0 O0 G1 & N0 \textless= -0.45 & 0.2002 & G1 \textless= -0.43 & 0.197 & O0 \textless= -0.43 & 0.0723 & 0.08 \textless G1 \textless= 0.59 & -0.0681\\ \hline
Positive M1 B0 J1 M0 & J1 \textless= -0.24 & 0.1488 & M0 \textless= -0.26 & 0.1306 & M1 \textless= -0.25 & 0.1281 & -0.24 \textless J1 \textless= 0.12 & -0.0685\\ \hline
High-Low T1 P1 S1 I1 & S1 \textless= -0.44 & 0.1492 & I1 \textless= -0.22 & 0.1203 & P1 \textless= -0.51 & 0.0571 & -0.44 \textless S1 \textless= 0.11 & -0.0528\\ \hline
Switch A1 O0 C1 L1 & C1 \textless= -0.49 & 0.1488 & A1 \textless= -0.46 & 0.1463 & -0.49 \textless C1 \textless= 0.05 & -0.0562 & O0 \textless= -0.43 & 0.0456\\ \hline
\end{tabular}
\end{table*}

\section{Conclusions and Future Work}
Using synthetic datasets with injected ground-truth patterns, this work evaluates RF surrogate models, PCA and LIME in terms of their ability to recover relevant features and provide interpretable representations of underlying structures. The results demonstrate that, although all three approaches can highlight important variables to some extent, none of them reliably identifies complete multi-feature patterns or clearly captures relationships among features.

The RF surrogate model showed strong and consistent ranking performance across pattern types and data dimensionalities, successfully prioritising relevant features. However, its outputs were limited to feature importance rankings, providing little insight into how features relate to one another or collectively form patterns. PCA exhibited strong performance when patterns aligned with dominant variance directions but degraded substantially as dimensionality increased or when patterns did not dominate global variance. Consequently, PCA proved sensitive to the statistical structure of the data and unreliable for detecting more complex or subtle patterns. LIME, while offering rule-based explanations and successfully identifying influential features, produced fragmented valid rules that did not coherently represent global feature relationships. Correlation-based and switching patterns, in particular, could not be directly recovered due to the method’s reliance on independent univariate conditions.

Taken together, these findings highlight a broader methodological limitation: current widely used techniques are primarily designed for feature ranking, dimensionality reduction, or local explanation rather than true pattern detection. As a result, identifying structured multi-feature behaviours remains challenging, even in controlled settings where patterns are known to exist. This limitation has important implications for downstream tasks such as cluster analysis and exploratory data analysis, where meaningful grouping often depends on detecting interactions and coordinated behaviour among features rather than assessing variables independently. The lack of dedicated pattern detection methodologies therefore represents a significant gap in current data analysis practice. Future work should focus on developing new approaches capable of explicitly identifying multi-feature patterns and modelling relationships among variables. Such methods should move beyond independent feature scoring toward representations that capture collective behaviour, structural dependencies, and interpretable pattern boundaries. 

Evaluation of existing pattern detection methods against real-world datasets was beyond the scope of this work. Future work should expand this evaluation to real-world datasets. Whilst testing on synthetic datasets provides a controlled experimental environment, evaluation of existing approaches' efficacy identifying known and documented patterns in real-world datasets would improve confidence in the evaluation. This would also provide a comparative baseline for future pattern detection methods to be evaluated against, further improving confidence in novel approaches. 

\section{Acknowledgements}
Benjamin Connor is supported by a Department for the Economy Research Studentship.

\bibliographystyle{IEEEtran}
\bibliography{bibligraphy}

@article{bandyapadhyay2023find,
  title={How to find a good explanation for clustering?},
  author={Bandyapadhyay, Sayan and Fomin, Fedor V and Golovach, Petr A and Lochet, William and Purohit, Nidhi and Simonov, Kirill},
  journal={Artificial Intelligence},
  volume={322},
  pages={103948},
  year={2023},
  publisher={Elsevier}
}

@inproceedings{moshkovitz2020explainable,
  title={Explainable k-means and k-medians clustering},
  author={Moshkovitz, Michal and Dasgupta, Sanjoy and Rashtchian, Cyrus and Frost, Nave},
  booktitle={International conference on machine learning},
  pages={7055--7065},
  year={2020},
  organization={PMLR}
}

@article{loftus2022phenotype,
  title={Phenotype clustering in health care: a narrative review for clinicians},
  author={Loftus, Tyler J and Shickel, Benjamin and Balch, Jeremy A and Tighe, Patrick J and Abbott, Kenneth L and Fazzone, Brian and Anderson, Erik M and Rozowsky, Jared and Ozrazgat-Baslanti, Tezcan and Ren, Yuanfang and others},
  journal={Frontiers in artificial intelligence},
  volume={5},
  pages={842306},
  year={2022},
  publisher={Frontiers Media SA}
}

@article{marin2024integrating,
  title={Integrating fuzzy C-means clustering and explainable AI for robust galaxy classification},
  author={Mar{\'\i}n D{\'\i}az, Gabriel and G{\'o}mez Medina, Raquel and Aij{\'o}n Jim{\'e}nez, Jos{\'e} Alberto},
  journal={Mathematics},
  volume={12},
  number={18},
  pages={2797},
  year={2024},
  publisher={MDPI}
}

@article{alvarez2024comprehensive,
  title={A comprehensive framework for explainable cluster analysis},
  author={Alvarez-Garcia, Miguel and Ibar-Alonso, Raquel and Arenas-Parra, Mar},
  journal={Information Sciences},
  volume={663},
  pages={120282},
  year={2024},
  publisher={Elsevier}
}

@article{yamga2023interpretable,
  title={Interpretable clinical phenotypes among patients hospitalized with COVID-19 using cluster analysis},
  author={Yamga, Eric and Mullie, Louis and Durand, Madeleine and Cadrin-Chenevert, Alexandre and Tang, An and Montagnon, Emmanuel and Chartrand-Lefebvre, Carl and Chass{\'e}, Micha{\"e}l},
  journal={Frontiers in Digital Health},
  volume={5},
  pages={1142822},
  year={2023},
  publisher={Frontiers Media SA}
}

@article{jain2010data,
  title={Data clustering: 50 years beyond K-means},
  author={Jain, Anil K},
  journal={Pattern recognition letters},
  volume={31},
  number={8},
  pages={651--666},
  year={2010},
  publisher={Elsevier}
}

@article{belle2021principles,
  title={Principles and practice of explainable machine learning},
  author={Belle, Vaishak and Papantonis, Ioannis},
  journal={Frontiers in big Data},
  volume={4},
  pages={688969},
  year={2021},
  publisher={Frontiers Media SA}
}

@article{werner2023explainable,
  title={Explainable hierarchical clustering for patient subtyping and risk prediction},
  author={Werner, Enrico and Clark, Jeffrey N and Hepburn, Alexander and Bhamber, Ranjeet S and Ambler, Michael and Bourdeaux, Christopher P and McWilliams, Christopher J and Santos-Rodriguez, Raul},
  journal={Experimental Biology and Medicine},
  volume={248},
  number={24},
  pages={2547--2559},
  year={2023},
  publisher={SAGE Publications Sage UK: London, England}
}

@article{dunn2018cluster,
  title={Cluster analysis in nursing research: an introduction, historical perspective, and future directions},
  author={Dunn, Heather and Quinn, Laurie and Corbridge, Susan J and Eldeirawi, Kamal and Kapella, Mary and Collins, Eileen G},
  journal={Western journal of nursing research},
  volume={40},
  number={11},
  pages={1658--1676},
  year={2018},
  publisher={Sage Publications Sage CA: Los Angeles, CA}
}

@inproceedings{ribeiro2016should,
  title={" Why should i trust you?" Explaining the predictions of any classifier},
  author={Ribeiro, Marco Tulio and Singh, Sameer and Guestrin, Carlos},
  booktitle={Proceedings of the 22nd ACM SIGKDD international conference on knowledge discovery and data mining},
  pages={1135--1144},
  year={2016}
}

@article{jolliffe2016principal,
  title={Principal component analysis: a review and recent developments},
  author={Jolliffe, Ian T and Cadima, Jorge},
  journal={Philosophical transactions of the royal society A: Mathematical, Physical and Engineering Sciences},
  volume={374},
  number={2065},
  pages={20150202},
  year={2016},
  publisher={the Royal Society publishing}
}

@article{luss2010clustering,
  title={Clustering and feature selection using sparse principal component analysis},
  author={Luss, Ronny and d’Aspremont, Alexandre},
  journal={Optimization and Engineering},
  volume={11},
  number={1},
  pages={145--157},
  year={2010},
  publisher={Springer}
}

@article{zafar2019dlime,
  title={DLIME: A deterministic local interpretable model-agnostic explanations approach for computer-aided diagnosis systems},
  author={Zafar, Muhammad Rehman and Khan, Naimul Mefraz},
  journal={arXiv preprint arXiv:1906.10263},
  year={2019}
}

\end{document}